\documentclass{vgtc}                          

\ifpdf
  \pdfoutput=1\relax                   
  \usepackage{graphicx}                
  \DeclareGraphicsExtensions{.pdf,.png,.jpg,.jpeg} 
\else
  \usepackage{graphicx}                
  \DeclareGraphicsExtensions{.eps}     
\fi%

\graphicspath{{figures/}{pictures/}{images/}{./}} 

\usepackage{microtype}                 
\PassOptionsToPackage{warn}{textcomp}  
\usepackage{textcomp}                  
\usepackage{mathptmx}                  
\usepackage{times}                     
\usepackage{cite}                      
\usepackage{tabu}                      
\usepackage{booktabs}                  
\usepackage{kotex}
\usepackage{comment}
\usepackage{xcolor}
\usepackage{xspace} 
\newcommand{\toolname}{SANVis\xspace}
\newcommand{\headlens}{HeadLens\xspace}

\onlineid{1189}

\vgtccategory{Research}

\vgtcinsertpkg

\teaser{
  \includegraphics[width=\textwidth]{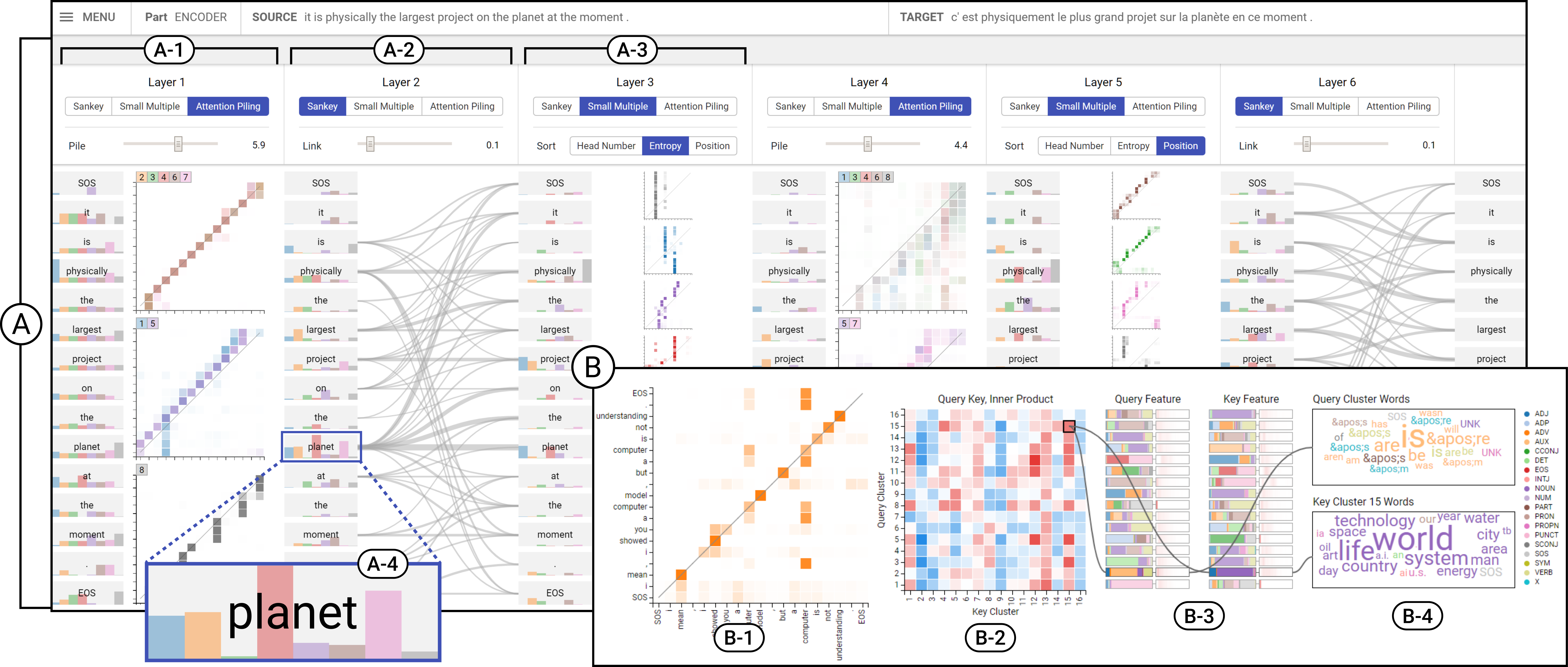}
  \caption{Overview of \toolname. (A) The network view displays multiple attention patterns for each layer according to three type of visualization options: (A-1) the attention piling option, (A-2) the Sankey diagram option, and (A-3) the small multiples option. (A-4) The bar chart shows the average attention weights for all heads (each colored with its corresponding hue) per each layer. 
  (B) The \headlens view helps the user analyze what the attention head learned by showing representative words and by providing statistical information of part-of-speech tags and positions.}
  \label{fig:teaser}
}

\newcommand{\domainlink}{\url{http://short.sanvis.org}\xspace}

\title{\toolname: Visual Analytics for Understanding Self-Attention Networks}

\author{
\begin{tabular}{ccccc}
Cheonbok Park\thanks{\{cb\_park, windy9898, vkfwlsdhs, jchoo\}@korea.ac.kr}
 & Inyoup Na\footnotemark[1]
 & Yongjang Jo\thanks{jyj3312@gmail.com} 
 & Sungbok Shin\thanks{sbshin90@cs.umd.edu}
 & Jaehyo Yoo\footnotemark[1] \\
 \scriptsize Korea University
 & \scriptsize Korea University
 & \scriptsize Korea University
 & \scriptsize University of Maryland
 & \scriptsize Korea University \\
 Bum Chul Kwon\thanks{bumchul.kwon@us.ibm.com}
 & Jian Zhao\thanks{jianzhao@uwaterloo.ca; work was completed while at FXPAL.}
 & Hyungjong Noh\thanks{\{nohhj0209,yeonsoo\}@ncsoft.com}
 & Yeonsoo Lee\footnotemark[6]
 & Jaegul Choo\footnotemark[1] \\
 \scriptsize IBM Research
 & \scriptsize University of Waterloo
 & \scriptsize NCSOFT Co., LTD.
 & \scriptsize NCSOFT Co., LTD.
 & \scriptsize Korea University
\end{tabular}
}

\abstract{Attention networks, a deep neural network architecture inspired by humans' attention mechanism, have seen significant success in image captioning, machine translation, and many other applications. 
Recently, they have been further evolved into an advanced approach called multi-head self-attention networks, which can encode a set of input vectors, e.g., word vectors in a sentence, into another set of vectors. Such encoding aims at simultaneously capturing diverse syntactic and semantic features within a set, each of which corresponds to a particular attention head, forming altogether multi-head attention. Meanwhile, the increased model complexity prevents users from easily understanding and manipulating the inner workings of models. To tackle the challenges, we present a visual analytics system called \toolname, 
which helps users understand the behaviors and the characteristics of multi-head self-attention networks. Using a state-of-the-art self-attention model called Transformer, we demonstrate usage scenarios of \toolname in machine translation tasks. Our system is available at \domainlink.
} 

\CCScatlist{
  \CCScatTwelve{Deep neural networks, visual analytics, natural language processing, interpretability, self-attention networks}{}{}{}
  
}

\begin{document}

\firstsection{Introduction}

\maketitle

Attention-based deep neural networks, inspired by humans' attention mechanism, are widely used for sequence-to-sequence modeling, e.g., neural machine translation
The attention module allows the model to dynamically utilize different parts of the input sequence, which leads to state-of-the-art performances in natural language processing (NLP) tasks~\cite{bahdnamu2014seq2attn,xu2015showattend,luong2015effective}. 

However, conventional approaches using recurrent neural networks (RNNs) had limitations that (1) they utilize only a single attention module that can capture only a particular characteristic of a given input and that (2) they cannot properly capture long-range dependencies due to the loss in memory content over multiple time steps. 


To address these limitations, Vaswani et al.~\cite{vasw2017transformer} recently proposed multi-head self-attention networks (in short, self-attention networks), which replace an RNN-based sequence encoding module with a sophisticated attention module. This module is composed of multiple different attention heads, each of which captures its own syntactic and/or semantic features within a set. Owing to these advantages, self-attention-based models have achieved state-of-the-art performances in machine translation, and it has been further extended in other NLP tasks~\cite{jacob2018bert,radford2019gpt2} and computer vision domains~\cite{Wang2018nonlocal,Zhang2018Sagan}. 

However, their highly sophisticated model architecture prevents users from deeply understanding and interacting with them. In response, this paper presents a visual analytics system for self-attention networks, called \toolname, as well as its comprehensive usage scenarios using widely-used networks called Transformer~\cite{vasw2017transformer}. 

\begin{figure}
    \centering
    
    \includegraphics[width=1\columnwidth]{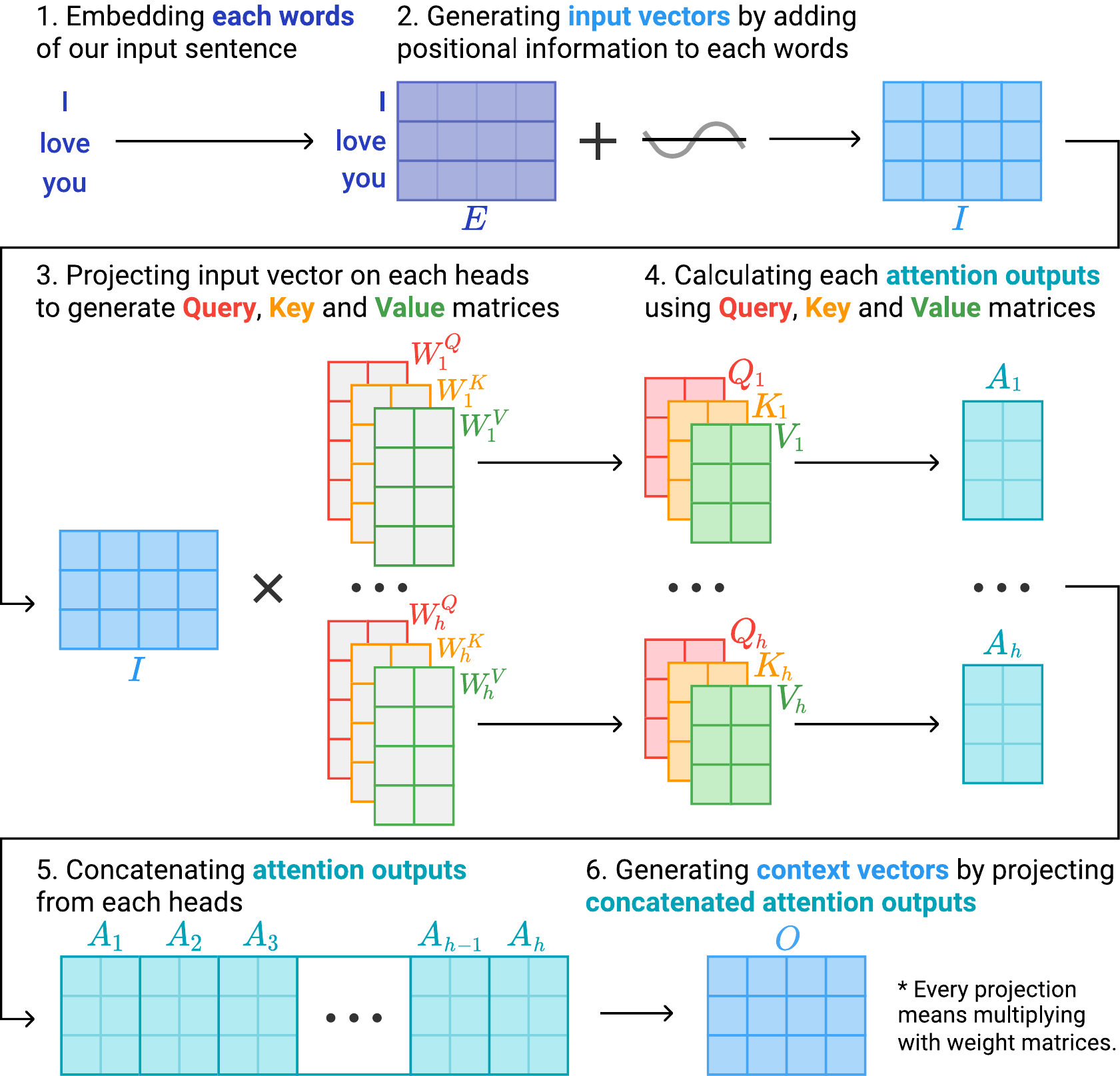}\vspace{-0.1in}
    \caption{How a multi-head self-attention module works. Steps 1 and 2 correspond to the embedding layer, while Steps 3 to 6 correspond to a single-layer multi-head self-attention module. 
    }
    \label{fig:background_model}\vspace{-0.2in}
\end{figure}

\section{Related Work}

We discuss related work from two perspectives: (1) visual analytic approaches for interpreting and interacting with various deep neural networks and (2) interpretation and analysis of self-attention networks mainly in NLP domains. 

Regarding the former, various visual analytic approaches have been proposed for convolutional neural networks mainly computer vision domains~\cite{liu16cnnvis,zeiler2014visualizing,Pezzotti17DeepEyes,Liu18DeepTracker,alsall2017hierachy,revacnnfilm} and RNNs in NLP domains~\cite{ming2017rnnvis,strobelt2018lstmvis,strobelt_seq2seq-vis_2019,cashman_rnnbow_2018,kwon_retainvis_2019}. Visual analytic approaches have also been integrated with other advanced neural network architectures, such as generative adversarial networks~\cite{kahng18GanLab,wang2018GANViz}, deep reinforcement learning~\cite{wang2019dqnvis}. Among them, Strobelt et al.~\cite{hendrik2018seq} developed a visual analytic system for RNN-based attention models, mainly for the exploration and understanding of sequence-to-sequence modeling tasks. However, despite the success of multi-head self-attention networks, such as BERT~\cite{jacob2018bert} and XLNet~\cite{Yang2019XLNetGA}, visual analytic approaches for these advanced attention networks have not existed before.

In NLP domains, recent studies~\cite{jacob2018bert,Vig2019AnalyzingTS,Clark2019WhatDB} have analyzed diverse behaviors of different attention heads in a self-attention model and have drawn linguistic interpretations as to what kind of syntactic and/or semantic features each attention head captures. Another line of research~\cite{Voita2019AnalyzingMS,Strubell2018LinguisticallyInformedSF} have attempted to leverage insights obtained from such in-depth analysis to improve the prediction accuracy and computational efficiency by, say, removing unnecessary heads and refining them. However, these approaches have not properly utilized the potential of interactive visualization, so in this respect, our work is one of the first sophisticated visual analytics systems for self-attention networks. 

\begin{figure}
    \centering
    \includegraphics[width=1\columnwidth]{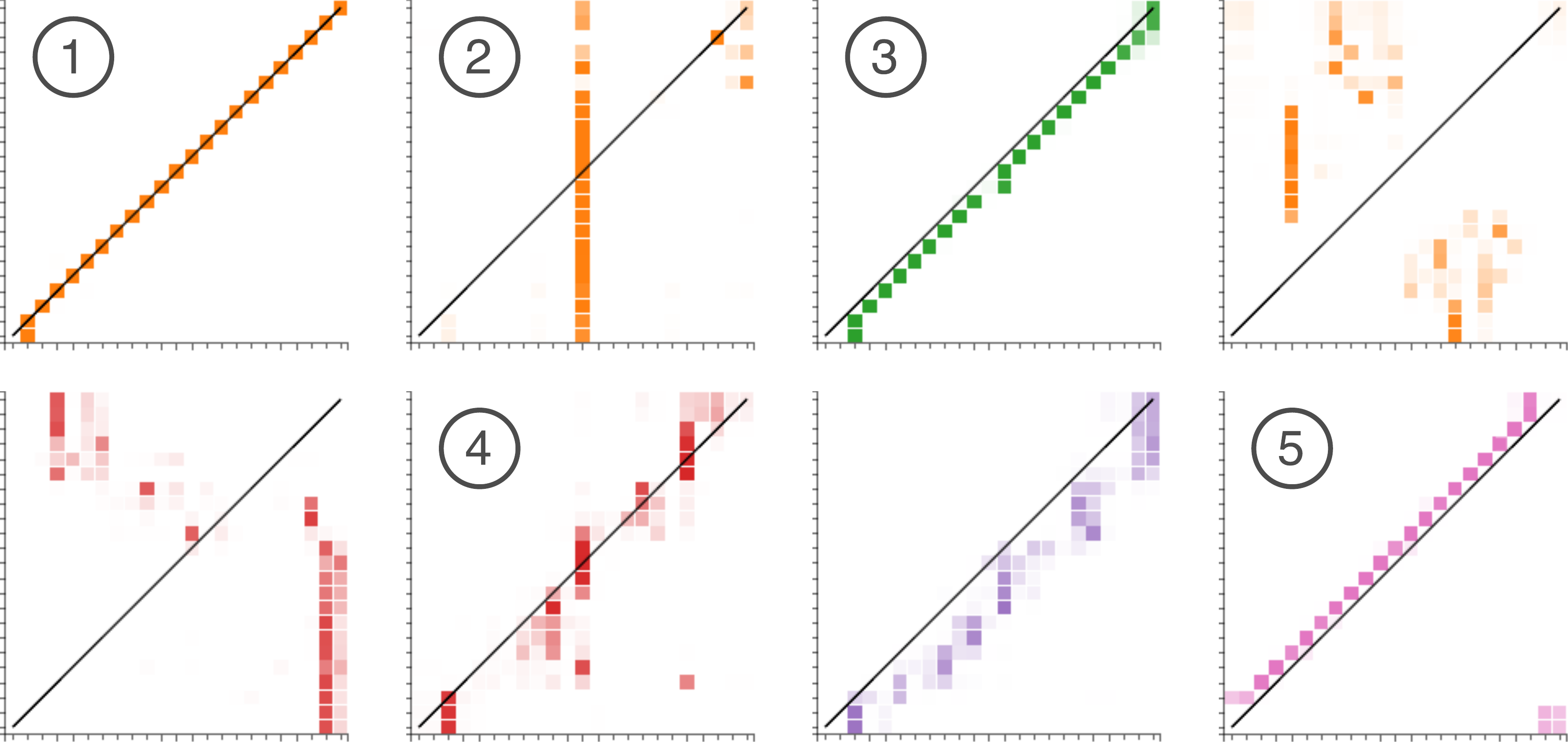}\vspace{-0.1in}
    \caption{Diverse attention patterns found in the encoder of Transformer. Some attention heads show diagonal patterns indicating that a query word attends to itself (1) or its immediate previous (5) or next word (3). Some other attention heads attend to a common, single word (2). In other attention heads, each group of consecutive words attends commonly to a single word within that group (4).}
    \label{fig:attention pattern}\vspace{-0.2in}
\end{figure}

\section{Preliminaries: Self-Attention Networks}
This section briefly reviews the self-attention module originally proposed in Transformer~\cite{vasw2017transformer}. 
Transformer adopts an encoder-decoder architecture to solve sequence-to-sequence learning tasks, e.g., neural machine translation, which converts a sentence in one language into that in another language.  
It converts a sequence of words in one domain into that in another domain. For example, for machine translation tasks, it translates a sentence in one language into that in another language.
In this process, the encoder of Transformer converts input words (e.g., English words) to internal, hidden-state vectors, and the decoder turns the vectors into a sequence of output words (e.g., French words).

Each encoder and decoder respectively consists of multiple layers of computing functions inside.
Furthermore, each layer in the encoder includes two sequential sub-layers, which are a multi-head self-attention and a position-wise feed-forward network. 
In addition to the multi-layer architecture of the encoder, the decoder has an additional attention layer, which is called as an encoder-decoder attention and helps the model to give attention to the encoders' internal states. 
Each layer of both encoder and decoder also consists of skip-connection and layer normalization in their computation pipeline. Overall encoder and decoder architecture are the stacks of $L$ identical encoder layers or decoder layers, including an embedding layer. 

We summarize the computation process with mathematical notations, so readers are advised to read the remaining section for details:
Let us denote $d_{model}$ as the size of hidden state vector and $h$ as the number of heads in multi-head self-attention. Each dimension of query, key, and value vector is $d_{q}=d_{k}=d_{v} = d_{model}/h$.

The embedding layer transforms the input token $x_{i}$ to its embedding space $e_{i}$ using a word embedding and adds the position information for each input token using sinusoidal functions (see Steps 1 and 2 in Figure~\ref{fig:background_model}), where $x_{i}$ is the $i$-th input token in $X=[x_1,\cdots,x_T]$. 




At each attention head, we transform encoded word vectors into three matrices of a query, a key, and a value, $Q \in R^{T \times d_q}$, $K \in R^{T \times d_k}$, and $V \in R^{T \times d_v}$, respectively, for $h$ times, which in turn generated $h$$\times$$3$ matrices, using the linear transformation and compute the attention-weighted combinations of value vectors as 
\begin{equation}\label{eqn:2}
\begin{array}{r@{}l}
\textrm{Attention}(Q,K,V) & = \textrm{Softmax}\left(\frac{QK^{T}}{\sqrt{d_{model}}}\right)V \\
\textrm{MultiHeadAttention} & =\textrm{Concat}\left(head_{1},\dots,head_{h}\right)W^O
\end{array}
\end{equation}
where $head_{i} = \textrm{Attention}\left(QW_{i}^{Q},KW_{i}^K,VW_{i}^{V}\right)$, and  $W_{i}^Q$, $W_{i}^V$ and $W_{i}^K$ indicate the linear transformation matrices at the $i$-th head. In multi-head self-attention, which consists of $h$ parallel attention heads, transformation  matrices of each head are randomly initialized, and then each set is used to  project input vectors onto a different representation subspace. For this reason, every attention head is allowed to have different attention shapes and patterns. This characteristic encourages each head differently to attend adjacent words or linguistics relation words. 


In the encoder layer, source words (input words to the encoder) work as the input to the query, key, and value transformations at the $i$-th head. In the decoder layer, the input can vary by attention types. While the decoders' self-attention takes target words (output words of the decoder) as its input, the encoder-decoder attention has target words as input to a query transformation but source words as the input to a key and a value transformation. 

\begin{figure}
    \includegraphics[width=1\columnwidth]{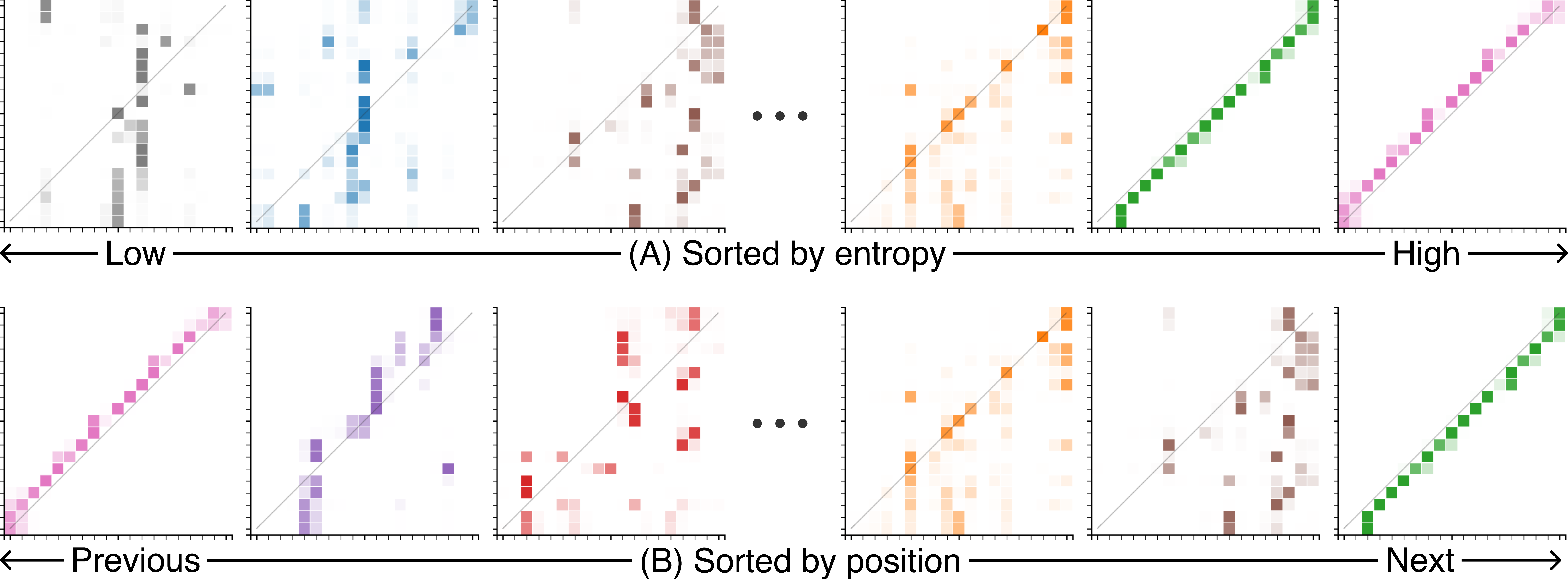}\vspace{-0.1in}
    \caption{Attention sorting result. The user can sort a set of multiple attention patterns with respect to different criteria such as the entropy measure (A) and the relative positional offset from query words (B).}
    \label{fig:attention_sort}\vspace{-0.2in}
\end{figure}

\section{Design Rationale}

We consider our design rationale of \toolname as follows: 

\noindent\textbf{R1: Grasping the information flow across multiple layers.} 

\noindent\textbf{R2: Identifying and making sense of attention patterns of each attention head.} 




\noindent\textbf{R3: Visualizing syntactic and semantic information to allow of exploring the attention head in their query and key vectors.}

\section{\toolname}
We present \toolname,\footnote{Our demo is available at \domainlink.} a visual analytics system for deep understanding of self-attention models, as shown in Figure~\ref{fig:teaser}. 
\toolname is composed of the network view and the \headlens view.
(1) The network view allows the user to understand the overall information flow through our visualization across the multiple layers (T1). Moreover, this view provides additional visualization options that assist the user in distinguishing distinct patterns from multiple attention patterns within a layer (T2). (2) The \headlens view reveals the characteristics of the query and the key vectors and their relationship of a particular head (T3).

\subsection{Network View}
Network view mainly visualizes the overview of attention patterns across multiple layers using the Sankey diagram (T1). Additionally, this view supports `piling' and `sorting' capabilities to understand common as well as distinct attention patterns among multiple attention heads (T2). For example, one can replace the Sankey diagram with a multiple heatmap view, where multiple heatmaps corresponding to different heads can be sorted by several different criteria (Figure,~\ref{fig:teaser} (A-3)). The attention piling view aggregates multiple attention patterns into a small number of clusters (Figure~\ref{fig:teaser}(A-1)). 

As shown in Figure~\ref{fig:teaser}(A), a set of words are sequentially aligned vertically in each layer, and represented the histogram according to attention weights from multiple heads. In Figure~\ref{fig:teaser}(A-4), each bar corresponds to a particular head within the layer where its height represents the total amount of attention weights assigned to those words by a specific head. As with Figure~\ref{fig:teaser}(A-2), if the fourth head in the layer attended to the word `planet' more highly than others, the fourth bar would be higher than the others. In this manner, the user easily recognizes which heads highly attend those words based on the height of histogram bars. Furthermore, when the user moves the mouse over the particular color bar, we show an attention heat map of the corresponding head in that layer.

\textbf{Sankey diagram.} As shown in Figure ~\ref{fig:teaser}(A-2), the edge weight between them represents the average attention weight across multiple heads within a particular layer. In this figure, we can see the strong link between `planet' in layer 2 to the preposition words(`on'), the same word and that article(`the') in layer 3. 
It means a significant amount of information of `planet' in layer 2 is conveyed to encode each word of a phrase (``on the planet'') in layer 3. This pattern shows the model captures the context meaning of the word, which is defined as the linguistic phrase in the given sentence, for improving the quality of translation.

\textbf{Attention Sorting.} Figure~\ref{fig:attention pattern} shows various attention patterns between query (y-axis) and key (x-axis) words for the different attention head in different layers, where a gray diagonal line indicates the position of attending itself for detecting attention patterns.

We focus on reducing the users' efforts to find the distinguish attention patterns by using sorting algorithms, which is based on relative positional information and the entropy value in the attention  (Figure~\ref{fig:attention pattern}). 
Relative positional information, such as whether the attention goes mainly toward the left, right, or the current location, as well as the column-wise mean entropy value of the attention matrix, allow the users to detect these patterns easily.

 Figure~\ref{fig:attention_sort} shows the sorted results of attention patterns based on our position or entropy sorting algorithms.
 When sorted by position, a number of attention is unambiguous that attention that inclines towards the past words are placed near the control panel at the top while those that lean towards the future words are placed relatively close to the bottom. When sorted by entropy, the uppermost attention has the lowest entropy and exhibits bar-shaped attention, which numerous query words attend the same word. At the bottom, the user can find that no more words focused on the same word. 


\begin{figure}
    \includegraphics[width=1\columnwidth]{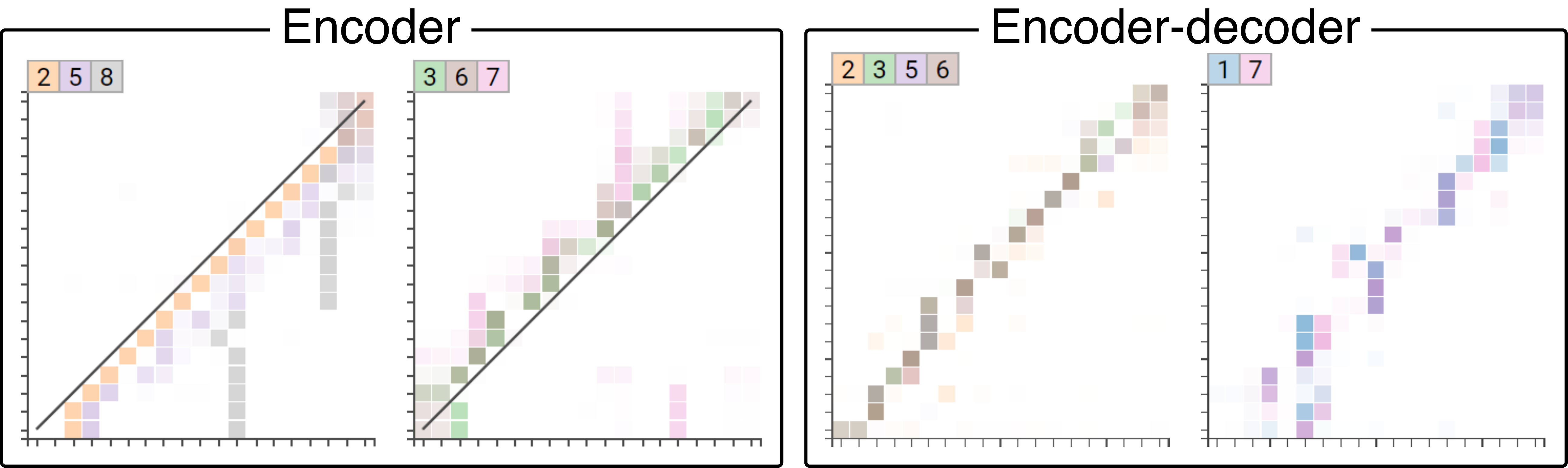}\vspace{-0.1in}
    \caption{Attention piling example in the encoder layer and encoder-decoder layer. In the encoder-decoder example, piling results do not have  a gray diagonal line because of the difference between the count of query words and key words.}
    \label{fig:attention_piling}\vspace{-0.2in}
\end{figure}

\textbf{Attention Piling.} Inspired by the heatmap piling methods~\cite{bach2015smallmultiples,Henry2007nodetrix}, we applied this piling idea to summarize multiple attention patterns in a single layer, as shown in the encoder part of Figure~\ref{fig:attention_piling}. To this end, we compute the feature vector of each attention head and perform clustering to the form of piles (or clusters) of multiple attention patterns. 

The feature vector of a particular attention on the attention head is defined as a flattened $n^2$-dimensional vector of its ${A_{i}} \in R^{T\times T}$ attention matrix, where ${A}_{i}$ is calculated from $\textrm{Softmax}\left(\frac{QK^{T}}{\sqrt{d_{model}}}\right)$ on the $i$-th head, concatenated with additional three-dimensional vector of (1) the sum of the upper triangular part of the matrix, (2) that of the lower-triangular part, and (3) the sum of diagonal entries. This three-dimensional vector  indicates the proportions how much attention is assigned to (1) the previous words of a query word, (2) its next words, (3) and the query itself, respectively. 

Using these feature vectors of multiple attention heads within a single layer, we perform hierarchical clustering based on their Euclidean distances. In this manner, multiple attention patterns are grouped, forming an aggregated heatmap visualization per computed pile along with head indices belonging to each pile, as shown in Figure~\ref{fig:attention_piling}. It helps the user to easily find the similar patterns and distinct patterns in the same layer by adjusting Euclidean distance.


\subsection{\headlens}

\begin{figure}[b]
    \vspace{-0.2in}
    \includegraphics[width=\columnwidth]{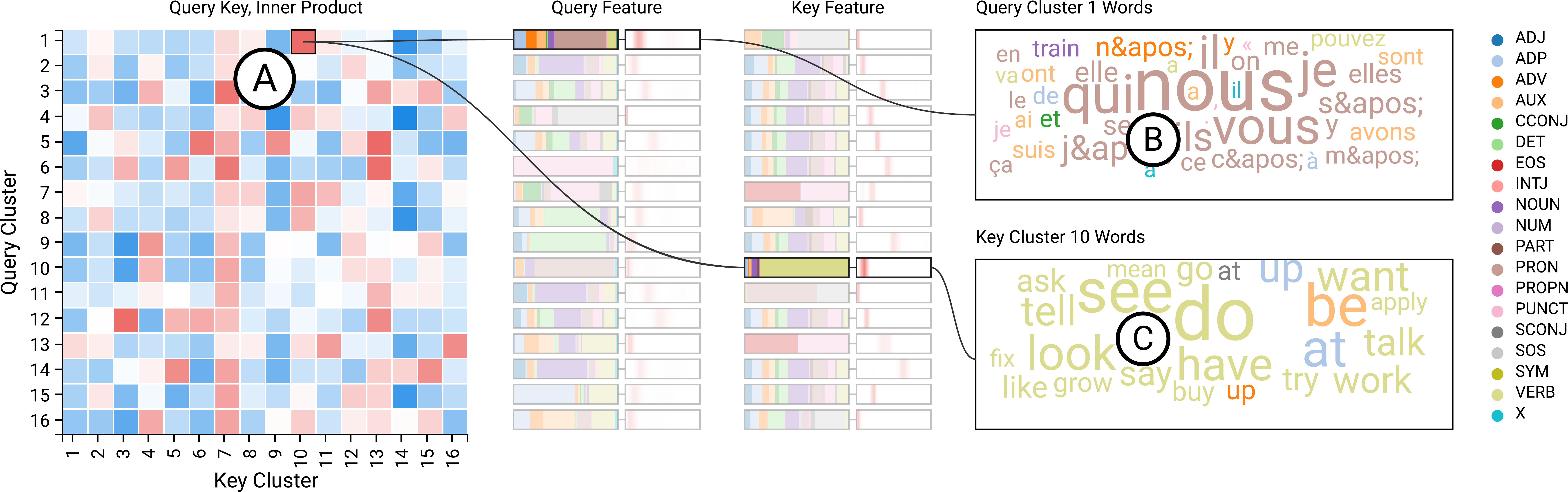}\vspace{-0.1in}
    \caption{\headlens showing the encoder-decoder attention of head 7 in layer 4. }
    \label{fig:enc_dec_lens}
\end{figure}
To analyze a particular attention head, \toolname offers a novel view called the \headlens, as shown in Figure~\ref{fig:teaser}(B). This view facilitates detailed analysis of the query and key representations of the selected attention head, such as which linguistic or positional feature they encoded (T3).  This view opens when a user clicks a particular heatmap corresponding to an attention head in the network view. 

The \headlens is generated as follows. (1) It performs clustering on query and key vectors separately. (2) For each pair of a query and a key cluster centroids, its pairwise similarity is computed, forming a heatmap visualization in Figur~\ref{fig:teaser} (B-2). (3) Additionally, the POS tagging and the positional information is summarized for each of the query and the key clusters (Figure~\ref{fig:teaser}(B-3)). (4) Once a user clicks a particular cell in a heatmap, its corresponding query and a key cluster are summarized in terms of their representative keywords (Figure~\ref{fig:teaser}(B-4)).

To be specific, in the first step, we consider all the sentences in a validation set and obtain the query and the key vectors of all the words from these sentences. These query and key vectors are the results of applying a query and a key transformation of input words for a given attention head. Next, we perform the $K$-means++~\cite{K_plus_mean} algorithm for each of the above-described query and key vector sets, by using the pre-defined number of clusters, e.g., 16 in our case. We empirically set this number of clusters by using an elbow method. 

In the second step, we obtain the cluster centroid vectors from the set of clusters for query vectors as well as those centroid vectors for key vectors. In the third step, we compute all the pairwise inner product similarities between each pair of a query cluster centroid and a key cluster one, which are visualized as a heatmap (Figure~\ref{fig:teaser} (B-2)). We choose the inner product as a similarity measure since the attention weight is mainly computed based on the inner product between a query and a key vector. Within each heatmap, the color of cells is marked red(or blue) if it has high(or low) similarity. High inner product between a query set and a key set means that the words in the query set are likely to attend to the words in the key set.

In the third step, the \headlens provides a summary of each of the query and the key clusters. Each query (or key) cluster contains those words whose query (or key) vectors belong to the cluster. For those words, we obtain their part-of-speech (POS) tags and position indices within the sentence which each of them appears in. For POS tags, we used universal POS tagger~\cite{stafordCorenlp}. Afterward, the relative amount of those words with each POS tag type out of the entire words within a single cluster is shown as a horizontal bar width with its encoded color, as shown in the left bar of Figure~\ref{fig:teaser} (B-3). In addition, the relative amount of those words shown in a particular position of their original sentences are color-coded (a higher value colored as a red), as shown in the right bar of Figure~\ref{fig:teaser} (B-3). 

Finally, in the fourth step, the user can click a particular entry in the cluster-level heatmap, e.g., a pair of a query and a key cluster with high similarity (a red cell highlighted in a black square in Figure~\ref{fig:teaser}(B-2)). Then, the summary of the corresponding query and key cluster are indicated by a black-colored edge (Figure~\ref{fig:teaser}(B-3)). Additionally, the word cloud visualization of such user-selected query and key cluster are used to highlight the frequently appearing words in each cluster, color-coded with their own POS tag types (Figure~\ref{fig:teaser}(B-4)). 

For example, the selected entry in Figure~\ref{fig:teaser}(B-2) indicates that the query cluster 15 has high similarity with the key cluster 15. The selected query cluster mainly contains auxiliary verb words (orange-colored), while the selected key cluster mainly contains noun words (purple-colored) in Figure~\ref{fig:teaser}(B-3). Furthermore, their most appearing words are shown in the word cloud view (Figure~\ref{fig:teaser}(B-3)), which means that this head assigns a high attention weight to these noun words, e.g., `world' and `life' when a query word is given as an auxiliary verb word, e.g., `is' and `are.' This result shows the selected head have captured the linguistic relationship of noun and auxiliary verb.

\section{Usage Scenarios}

This section demonstrates usage scenarios of \toolname, mainly focusing on the recently proposed Transformer. 
This model has shown superior performances in machine translation tasks, including English-French and English-German translation tasks in the WMT challenger~\cite{noauthor_acl_nodate}. Our implementation of the Transformer is based on the annotated Transformer~\cite{opennmt}. Our model parameter setting followed the base model in the original paper~\cite{vasw2017transformer}. We set our target task as English-French translation, where the collection of the scripts from TED lectures is used as our dataset~\cite{paul2018personalnmt}. The BLEU score of our model is shown as 38.4, which validates a reasonable level of performance. For evaluating our system, we used the validation set, which is not seen during training. 

\textbf{Attention Piling.} 
In Figure~\ref{fig:attention_piling}, the encoder-decoder part shows the attention piling visualization in encoder-decoder attention. In this example, one can observe that a number of attention heads have a diagonal attention pattern. 
An appropriate explanation of this diagonal shape would be that the words in French and English are generally aligned in the same order~\cite{bahdnamu2014seq2attn}.  For the debugging purpose, it proves that the model properly learned a linguistic alignment between the source sequence and the corresponding target sequence.

\textbf{\headlens.}
In the earlier example, we saw that most attention patterns between the English and the French words are diagonally-shaped between English and French words. One can analyze this pattern in detail by using our \headlens. As shown in Figure~\ref{fig:enc_dec_lens}, we chose head 4 in layer 7, which has such a diagonal attention pattern, and applied the \headlens. Once selecting the query and the key cluster pair with a high similarity (Figure~\ref{fig:enc_dec_lens} (A)), it is shown that the query clusters commonly have an pronoun as a dominant POS tag type (brown-colored in Figure~\ref{fig:enc_dec_lens} (B)). Most of query cluster words are subject words in French, for instance, `nous' and `vous' mean `we' and `you' in English, respectively. The corresponding key clusters' representative words are mostly verbs. This result demonstrates that the model attends verb words to predict verb tokens for translating from English to French, when the input token is subject.

\section{Conclusions}
In this paper, we present \toolname, a visual analytics system for self-attention networks, which supports in-depth understanding of multi-head self-attention networks at different levels of granularity. 
Using several usage scenarios, we demonstrate that our system provides the user with a deep understanding of the multi-head self-attention model in machine translation. 

As future work, we plan to extend our \headlens to perform clustering of value vectors. We evaluate our system by various researchers who use the multi-head self-attention networks. We also apply our method in other state-of-the-art self-attention based models, such as BERT~\cite{jacob2018bert} and  XLNet~\cite{Yang2019XLNetGA}.

\acknowledgments{
The authors wish to thank all reviewers who provided constructive feedback for our project.
This work was partially supported by NCSOFT NLP Center. This work was also supported by the National Research Foundation of Korea (NRF) grant funded bythe Korean government (MSIP) (No. NRF-2018M3E3A1057305).
}

\bibliographystyle{abbrv-doi}

\bibliography{main}
\end{document}